# Rain-Code Fusion : Code-to-code ConvLSTM Forecasting Spatiotemporal Precipitation

Takato Yasuno[1] Akira Ishii[1] Masazumi Amakata[1]

[1] Research Institute for Infrastructure Paradigm Shift, Yachiyo Engineering Co., Ltd., Asakusabashi 5-20-8, Taito-ku, Tokyo, Japan
{tk-yasuno,akri-ishii,amakata}@yachiyo-eng.co.jp

**Abstract.** Recently, flood damage has become a social problem owing to unexperienced weather conditions arising from climate change. An immediate response to heavy rain is important for the mitigation of economic losses and also for rapid recovery. Spatiotemporal precipitation forecasts may enhance the accuracy of dam inflow prediction, more than 6 hours forward for flood damage mitigation. However, the ordinary ConvLSTM has the limitation of predictable range more than 3-timesteps in real-world precipitation forecasting owing to the irreducible bias between target prediction and ground-truth value. This paper proposes a rain-code approach for spatiotemporal precipitation code-to-code forecasting. We propose a novel rainy feature that represents a temporal rainy process using multi-frame fusion for the timestep reduction. We perform rain-code studies with various term ranges based on the standard ConvLSTM. We applied to a dam region within the Japanese rainy term hourly precipitation data, under 2006 to 2019 approximately 127 thousands hours, every year from May to October. We apply the radar analysis hourly data on the central broader region with an area of $136 \times 148$ km$^2$. Finally we have provided sensitivity studies between the rain-code size and hourly accuracy within the several forecasting range.

## 1 Introduction

### 1.1 Rain-Code Fusion Based Precipitation Forecasting

For the past decade, flood damage has become a social problem owing to unexperienced weather conditions arising from climate change. An immediate response to extreme rain and flood situation is important for the mitigation of casualties and economic losses and for faster recovery. The spatiotemporal precipitation forecast may influence the accuracy of dam inflow prediction more than 6 hours forward for flood damage mitigation. In particular, in Japan, forecasting 6 hours forward is critical and legally requested for the basic policy strengthening flood control function of an existing dam [1]. To prevent social losses due to incoming heavy rain and typhoons, we attempt to predict flood situations at least 6 hours beforehand. The dam manager can inform downstream residents of the hazardous flood scenario via an announcement. Then, the elderly and children can escape to safety facilities. Therefore, forecasting precipitation 6 hours forward is crucial for mitigating flood damage and ensuring safety of people downstream. This paper proposes a rain-code approach for spatiotemporal precipitation forecasting. We propose a novel rainy feature fusion that represents a temporal rainy process including hourly multi-frames. We perform rain-code studies with various ranges based on spatiotemporal precipitation forecasting using the ConvLSTM. We applied to a dam region within the Japanese rainy term hourly precipitation data, under 2006 to 2019 approximately 127 thousands hours, every year from May to October. We can use the radar analysis hourly data on the central broader region with an area of $136 \times 148$ km$^2$, based on "rain-code" fusion that contains hourly multi-frames such as 3, 4, 6, and 12.

### 1.2 Related Works and Papers

**Deep Learning Methods for Water Resources.** Sit et al. [2] provided a comprehensive review, focusing on the application of deep-learning methods to the water sector, e.g., for monitoring, management, and governance of water resources. Their study provides guidance for the utilisation of deep-learning methods for water resources. There were 315 articles published from 2018 to the end of March2020, excluding editorials, and review papers. Surprisingly, among the reviewed papers, convolutional neural networks (CNNs) and long short-term memory networks (LSTMs) were the most widely investigated architectures. They pointed out their success of respective task in matrix prediction and sequence prediction, which are important in hydrologic modelling. In the field of weather forecasting, the CNNs and LSTMs are also used for quantitative precipitation estimation. Because the prediction of precipitation is generally a time-series problem, LSTM is commonly used. Wu et al. [3] designed a model combining a CNN and LSTM to improve the quantitative precipitation estimation accuracy. The proposed model uses multimodal data, including satellite data, rain gauge observations, and thermal infrared images. The CNN-LSTM model outperforms comparative models, such as the CNN, LSTM, and Multi-Layer Perceptron. Yan et al. [4] used a CNN model together with radar reflectance images to forecast the short-term precipitation for a local area in China. As a dataset, the radar reflection images and the corresponding precipitation values for one hour were collected. The model takes the images as inputs and returns the forecast value for one hour precipitation. Chen et al. [5] focused on precipitation nowcasting using a ConvLSTM model. The model uses radar echo data to forecast 30 or 60 minutes of precipitation. Thus, in the field of weather forecasting, CNNs and LSTM s have mainly been used for short-term precipitation "nowcasting," with a limited forecasting range of 30–60 minutes.

**Spatiotemporal Sequence Precipitation Forecasting.** Shi et al. surveyed the forecasting multi-step future of spatiotemporal systems according to past observations. They summarised and compared many machine-learning studies from a unified perspective [6]. These methods are classified two categories: classical methods and deep-learning methods. The former methods are subdivided into the feature-based methods, state-space models, and Gaussian processes. The latter methods include the deep temporal generative models, feedforward neural networks (FNNs), and recurrent neural networks (RNNs). Remarkable methods for spatiotemporal precipitation forecasting have been developed, e.g., ConvLSTM [7], TrajGRU [8], and PredRNN [9, 10]. Although the state-of-art method is applicable to the benchmark moving MNIST and regularised generated movies, real-world weather prediction is limited to short-range forecasting (two or three timesteps) owing to the chaotic phenomena, non-stationary measurements, unexperienced anomaly weather even if we use the-state-of-the-art architectures.

In contrast, in the field of urban computing, e.g., air-quality forecasting [11] and traffic crowd prediction [12], the CNN and LSTM architectures allow 48- to 96-hours forecasting and peak hour demand forecasting [13]. However, for precipitation forecasting in the case of uncertain extreme weather, it is difficult to accurately predict more than 6 hours longer range forecasting beyond short-term nowcasting, for example 30, 60 minutes, and 2 hours [5, 14]. Especially in Japan, the 6 hours forecasting dam inflow are required practically [1]. To mitigate the flood damage, it is critical to extend the predictability of precipitation forecasting for predicting the dam inflow and controlling the outflow to the downstream region.

### 1.3 Stretch Predictability Using Code-to-code Forecasting

Two problems are encountered in training the RNN algorithm: vanishing and exploding gradients [14]. An RNN model was proposed in the 1980s for modelling time series [15]. They suggested that we allow connections among hidden units associated with a time delay. Through these connections, the model can retain information about the past inputs, allowing it to discover temporal correlations between events that are possibly far from each other in the data (a crucial property for proper learning of time series) [16]. This crucial property of time series is similarly fitted to the property of the spatiotemporal sequence. In contrast to a simple location-fixed regular grid, the irregular moving coordinate, which is more complex, makes it more difficult to learn the information from the past inputs. In practice, standard RNNs are unable to store information about historical inputs for a long time, and the ability of the network to model the long-range dependence is limited [17]. Of course, numerous revised network strategies have been proposed, such as the stacked LSTM [18], memory cell with peephole connections, and the composite model to copy the weights of hidden-layer connections [19]. However, it is not yet completely solved to forecast long-range future outputs from the previous spatiotemporal inputs. This paper proposes a new insight "rain-code" feature based on the multi-frame inputs using a standard ConvLSTM algorithm. We present training results for rain-code based precipitation forecasting using real-world spatiotemporal data.

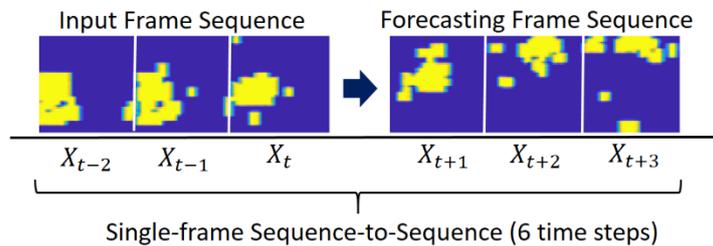

**Fig. 1.** Ordinary single-frame unit input and output for precipitation forecasting (This is the baseline model setting as the state-of-the-art **"Sequence-to-sequence"** ConvLSTM).

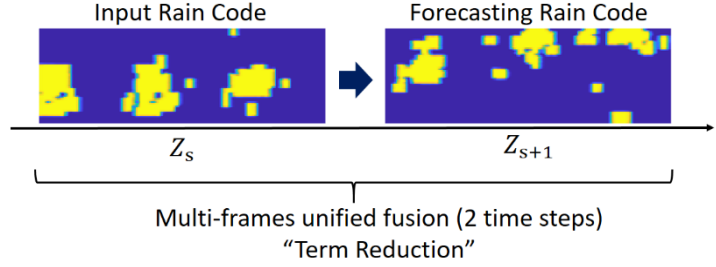

Fig. 2. Proposed rain-code based multi-frame input and output for precipitation forecasting (This is a proposed model setting **"Code-to-code"** ConvLSTM as general insight). This multi-frame composition per image results in the time step reduction for precipitation forecasting.

## 2  Precipitation Forecasting Method

### 2.1  Multi-Frame Rain-Code against Single-frame Sequence

**Rain-Code Fusion of Multi-Frame Feature.** Figure.1 shows an ordinary single-frame based input feature and target output sequence for precipitation forecasting. For example, in case of an hourly unit sequence, the left-side images show the sequence for the past 3 hours. The right-side images show the forward 3-hours target sequence. This single-frame unit window contains the total 6-hour timesteps. However, the ordinary ConvLSTM has the limitation of predictable range more than 3-timesteps in real-world precipitation forecasting, not simulated dataset as the MovingMNIST [7]. In contrast, Figure. 2 shows a proposed multi-frame-based input rain-code and target output rain-code for longer-term precipitation forecasting. In case of a 3-hours multi-frame rain-code, the left-side rainy process feature is unified by the past 3-hours 3-frame rain-code. The right-side process fusion is composed of the forward 3-hours target 3-frame rain-code. From a new fusion, this multi-frame window contains only two rain-codes. A new dataset for the rain-code based ConvLSTM is created with a narrower multi-frame sliding window, it results in the forecasting term reduction from 6 to 2. The rain-code based training allows rainy process features with temporal 3 hours to be learned simultaneously. This rain-code enable to represent the spatiotemporal relation with pair of 1-2 hour, 2-3 hour, and 1-3 hour. When we predict 2-timestep forecasting using the three-frame rain-code, 6-hours-forward precipitation forecasting is feasible.

### 2.2  Multi-Frame-Based Forecasting Precipitation Method

**Two-Timestep Forecasting Using Code-to-code ConvLSTM.** This paper proposes a rain-code based spatiotemporal precipitation forecasting. We introduce a new rainy feature fusion that represents a spatiotemporal rainy process including hourly multi-frames. We perform multi-frame studies with various ranges based spatiotemporal precipitation forecasting using the standard ConvLSTM [7]. We build the 2-layer LSTM networks, and the Rectified Linear Unit (ReLU) activation function is used in each post-layer, as shown Figure.3. The target size is 80 × 80 pixels, and the format is grayscale.

The filter size of the hidden layer is 80. The loss function is the cross entropy. The gradient optimiser is RMSProp, and the weight of square gradient decay 0.95. The learning rate is 0.001. At the first timestep, the ConvLSTM layer uses the inputs of initialized hidden layer and memory cell, respectively. At the next timestep, the ConvLSTM layer computes with the inputs of previous hidden layer and memory cell, recurrently. Here, we note that batch-normalization is not effective though we tried it. In this study, we confirmed that it is feasible to predict the two-timestep forecasting precipitation dataset.

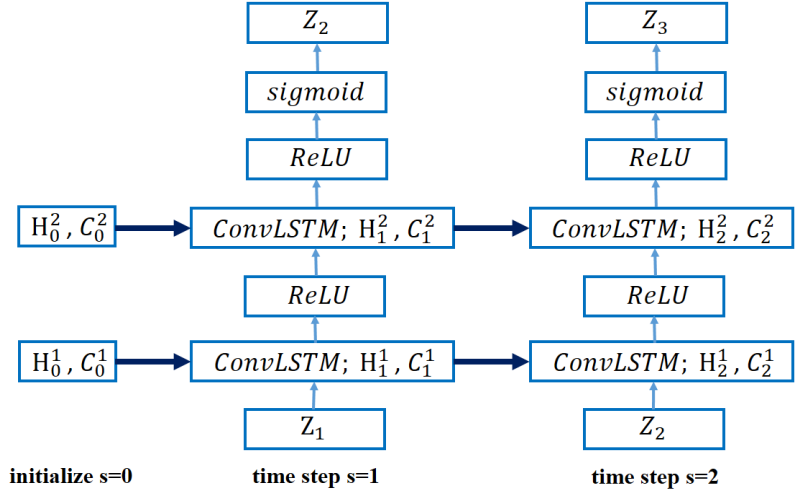

**Fig. 3.** Two-timestep Code-to-code ConvLSTM architecture using Rain-Code inputs and outputs.

### 2.3 Hourly Accuracy Indices for Rain-Code Forecasting

**Rain-Code Multi-Frame Accuracy.** We compute the root-mean-square error (RMSE) and the mean absolute error (MAE) of each 1- and 2-timestep rain-code based forecasting results using test data. These two indices are represented as follows.

$$\text{RMSE} = \frac{1}{SN^2}\sum_{s=1}^{S}\sum_{i=1}^{N}\sum_{j=1}^{N}\left[Z_s(i,j) - \hat{Z}_s(i,j)\right]^2 \quad (1)$$

$$\text{MAE} = \frac{1}{SN^2}\sum_{s=1}^{S}\sum_{i=1}^{N}\sum_{j=1}^{N}|Z_s(i,j) - \hat{Z}_s(i,j)| \quad (2)$$

Here, s stands for the timestep of rain-code, and $Z_s(i,j)$ indicates the ground truth of rain-code at timestep s, whose pixel location is the i-th row and j-th column. In contrast $\hat{Z}_s(i,j)$ denotes the predicted rain-code at timestep s, whose pixel location is $(i,j)$.

**Hourly Frame Accuracy.** We compute each hourly frame based RMSE and MAE forecasting results using test data. These two indices are formulated as follows.

$$\text{RMSE}(h) = \frac{1}{SK^2}\sum_{s=1}^{S}\sum_{k=1}^{K}\sum_{m=1}^{K}\left[F_s^h(k,m) - \hat{F}_s^h(k,m)\right]^2, h = 1,\dots,H. \quad (3)$$

$$\text{MAE}(h) = \frac{1}{SK^2}\sum_{s=1}^{S}\sum_{k=1}^{K}\sum_{m=1}^{K}|F_s^h(k,m) - \hat{F}_s^h(k,m)|, h = 1,\dots,H. \quad (4)$$

Here, h stands for the hourly-frame within s-th rain-code, and $F_s^h(k,m)$ indicates the ground truth of h-th frame at rain-code with timestep s, whose pixel location is the k-th row and m-th column. On the other hand, $F_s^h(k,m)$ denotes the predicted hourly frame feature at rain-code with timestep s, whose pixel location is $(k,m)$.

## 3 Applied Results

### 3.1 Training and Test Datasets for Matrix Prediction

We used the open-source Radar/Rain gauge Analysed Precipitation data [20], which have been provided by the Japan Meteorological Agency since 2006. They are updated every 10 minutes, with a spatial resolution of 1 km. The past data with a time interval of 30 minutes were displayed and downloaded. Figure 3 shows a display example of a real-time flood risk map for 15:10 JST, 30 September 2020. Additionally, the forecasts of hourly precipitation up to 6 hours ahead are updated every 10 minutes, with a spatial resolution of 1 km. Forecasts of hourly precipitation from 7 to 15 hours ahead are updated every hour, with a spatial resolution of 5 km.

We focused on one of the dams in the Kanto region with an area of approximately 100 km$^2$. We applied our method to this dam region of hourly precipitation data, under 2006 to 2019 approximately 127 thousands hours, every year from May to October whose term is the frequently rainy season in Japan. We downloaded and cropped the Radar Analysed Precipitation data on the dam central broader region with an area of $136 \times 148$ km$^2$, whose location is near the bottom left-side of Mount Fuji, as shown in Figure. 4. We aggregated $20 \times 20$ (400) grids for the beneficial feature enhancement.

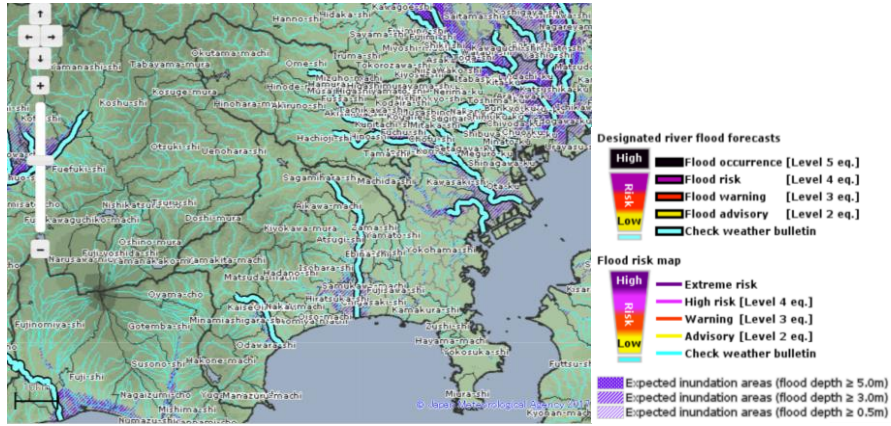

**Fig. 4.** Case study region in Kanto and an example of the real-time flood risk map (https://www.jma.go.jp/en/suigaimesh/flood.html).

**Table 1.** Percentile of hourly precipitation from 2006 to 2019 in the $136 \times 148$ km$^2$ region, whose area was aggregated by a sub-group of $20 \times 20$ grids.

| Percentile | Grid precipitation | Percentile | Grid precipitation |
|---|---|---|---|
| 82% | 0.008mm/hr | 99.9% | 15.25mm/hr |
| 85% | 0.10mm/hr | 99.99% | 32.18mm/hr |
| 90% | 0.40mm/hr | 99.999% | 50.65mm/hr |
| 95% | 0.92mm/hr | 99.9999% | 71.52mm/hr |
| 97% | 1.89mm/hr | 99.99999% | 93.80mm/hr |
| 99% | 4.70mm/hr | Maximum | 106.37mm/hr |

Table 1 presents the percentile of hourly precipitation from 2006 to 2019 (a total of 14 years) in the study region with an area of $136 \times 148$ km$^2$. To enhance the precipitation intensity, we aggregated the 1-km unit mesh into a subgroup of $20 \times 20$ grids (i, j = 1, ... ,20). For 81% of the 14 years, there was no precipitation; thus, this precipitation dataset was sparse. We excluded the empty of rainy term containing a zero-precipitation sequence, which was always 12 hours long. We transformed the mini-max form $B_{ij} = (A_{ij} - \min\{A_{ij}\})/(max\{A_{ij}\} - \min\{A_{ij}\})$ so that it ranged from 0 to 1. To efficiently learn the model, we excluded almost zero precipitation always 12 hours, whose grid precipitation is less than the threshold 1E+6 pixel counts. We set several thresholds and prepared a grayscale mask of dataset candidates; thus, we selected a practical level. Therefore, the rain-code dataset based on three frames contains the number of 37 thousands of the input and output features. The training data window includes three-frames based rain-codes ranging from 2006 May to 2018 October. The test dataset windows contains the last 30 windows with a 3-hours input rain-code and 3-hours output rain-code, whose 6 frames are multiplied by the 30 rain-code sets to obtain a 180-hours test set that stands for 7.5 days until 2019 October.

### 3.2 Feasibility Study of Rain-Code Forecasting and Computing Accuracy

We implemented the training for the 3-hours $1 \times 3$ frame-based rain-code. We set the number of mini-batches as 24 and performed the training for 10 epochs (approximately 1300 iterations). After 2.5 hours, the loss reached a stable level. Using this trained ConvLSTM network, we computed RMSE using equation (3) at each frame.

Figure 5 shows a plot of hourly frame based accuracy RMSE of the 1- and 2-timestep rain-code based forecasting test results. Here, the number of test rain-code cases was 12. Each RMSE was computed in the range of 0–8, and the average RMSE among the 12 cases was approximately 4. Both an increasing error and a decreasing error occurred, because the test rain-code had variations in the precipitation region. The patterns of the rain-code features (precipitation process) within 3 hours were classified, e.g., thin rain, decreasing rain, mid rain, increasing rain, and heavy rain. The prediction error was smaller for less rainy regions. The average frame based RMSEs forecasting results were almost constant level.

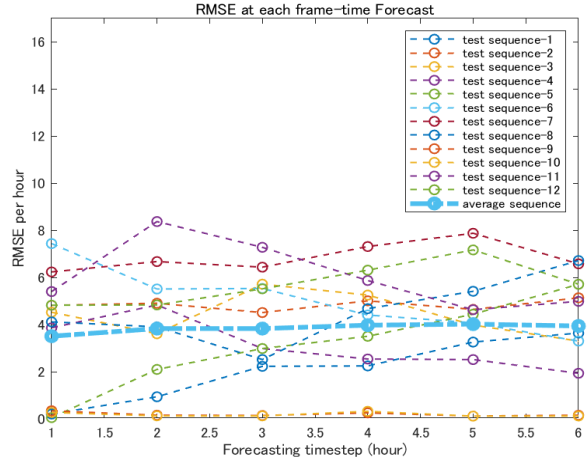

**Fig. 5.** Plot of the hourly frame based accuracy RMSE for the 1- and 2-timestep forecasting tests based on the 3-hours rain-code.

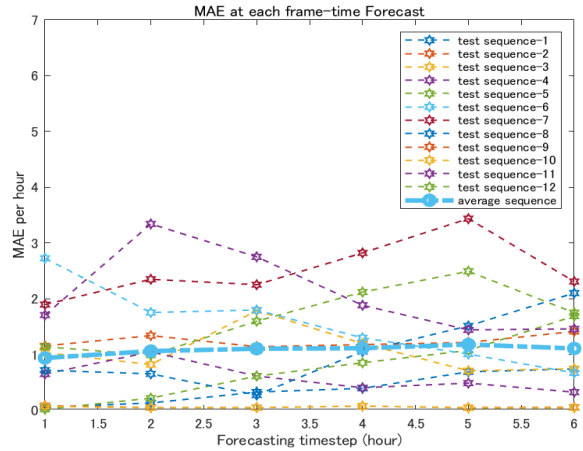

**Fig. 6.** Plot of the hourly frame based accuracy MAE for the 1- and 2-timestep forecasting tests based on the 3-hours rain-code.

We computed MAE using equation (4) at each frame. Figure 6 shows a plot of the frame based MAE of the 1- and 2-timestep rain-code based forecasting test results. Each MAE was computed in the range of 0–3.5, and the average MAE among the 12 cases was approximately 1. Both an increasing error and a decreasing error occurred, owing to the variations in the precipitation region for each test rain-code. Similar to the RMSE results, the prediction error was smaller for less rainy regions. The average frame based MAEs for the 1- and 2-timestep forecasting results were almost constant level. Therefore, both the RMSE and MAE were stable.

## 4    Rain-Code Size Sensitivity Studies

### 4.1    Rain-Code Training Results and Accuracy Indices for Different Number of Frames

We computed the RMSE and MAE using equation (1)(2) as the rain-code multi-frame based accuracy. Table 2 presents several training results and prediction error indices for different number of frames of the rain-code. We all trained the unified target size of $80 \times 80$; if the original rain-code size was different from the target size, we resized the rain-code to the unified size. However, we computed the prediction error indices resized back to the initial rain-code pixel size. Compared with four cases, the size of $2 \times 2$ frames rain-code outperform accuracy indices with both RMSE and MAE. This is because the rain-code was square rather than rectangular (as in the other cases); thus, the resized unified size operation and back to the original size influences a little information loss of rainy feature. A larger multi-frame size corresponded to a larger error of the rain-code based forecasting results. Unfortunately, a single-frame method as the baseline of Convolutional LSTM was impossible to predict the target 6 hours precipitation forecasting, as far as we implemented it to the dataset of study region from 2009 to 2016.

**Table 2.** Training results and indices with different multi-frame sizes of the of rain-code.

| Input Rain-Code multi-frame size | Target 6 hours Predictability | Average RMSE(h) | Average MAE(h) | Training time |
|---|---|---|---|---|
| Single-frame **Baseline** | 6 hours Not predictable | — | — | — |
| 3 (1×3frames) | 6 hours Predictable | 3.87 | 1.05 | 2hours 24min. (10 epoch) |
| **4 (2×2frames)** | **8 hours Predictable** | **3.68** | **1.02** | 1hours 52min. (10 epoch) |
| 6 (2×3frames) | 12 hours Predictable | 4.45 | 1.25 | 1hours 14min. (10 epoch) |
| 12(3×4frames) | 24 hours Predictable | 6.96 | 2.43 | 1hours 12min. (20 epoch) |

### 4.2    Rain-Code Based Forecasting Results and Accuracy Using 2×2 frames

Figure. 7 shows five test sets of prediction results at each line that contains the "ground truth" of rain-code (the left-side of two images) and the 2 time step rain-code based "forecasting result" (the right-side of two images), where each image is a rain-code feature composed by 4 hours range of $2 \times 2$ frames. This square frame size is the most accurate case that outperform RMSE and MAE than other rectangle shape of multi-frame size, shown as Table. 2.

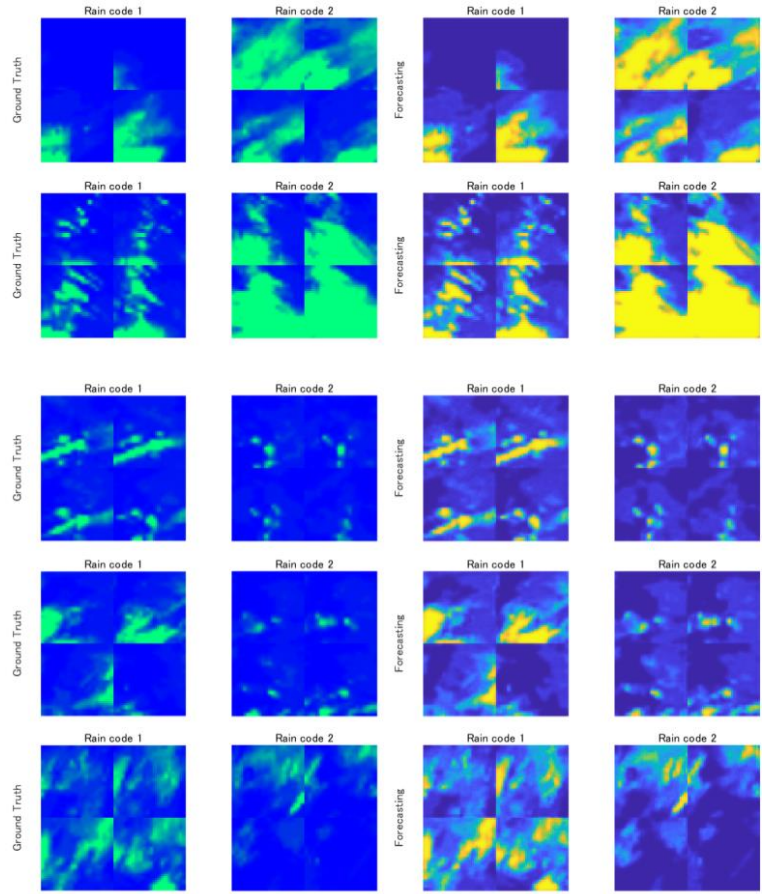

**Fig. 7.** Ground truth (left-side of two images) and 2 time step forecasting results (right-side of two images) using 2×2 frames 4 hours rain-code.

Figure. 8 shows a Plot of root mean square error (RMSE) of 1- and 2-time step rain-code based forecasting test results. Here, the number of test rain-code is 12 cases. Each RMSE is computed into the range from 0 to 10, and the average RMSE among 12 cases is around 3.5 level. In addition, Figure. 9 shows a Plot of mean absolute error (MAE) of 1- and 2-time step rain-code based forecasting test results. Each MAE is computed into the range from 0 to 7.5, and the average MAE among 12 cases is around 2 level.

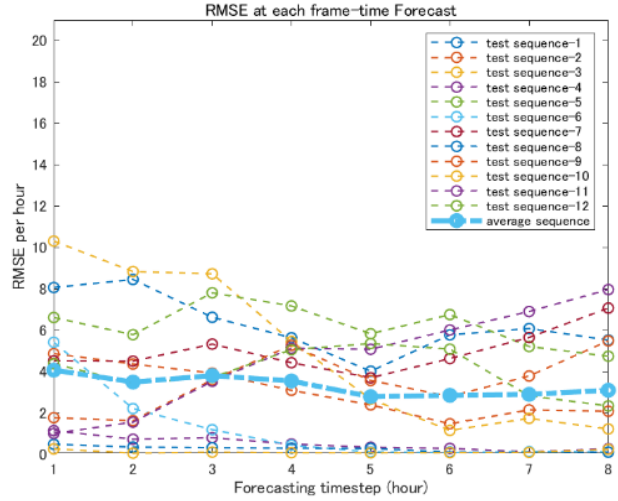

**Fig. 8.** Plot of RMSE hourly accuracy of 1- and 2-time step forecasting tests based on 2×2 frames 4 hours rain-code.

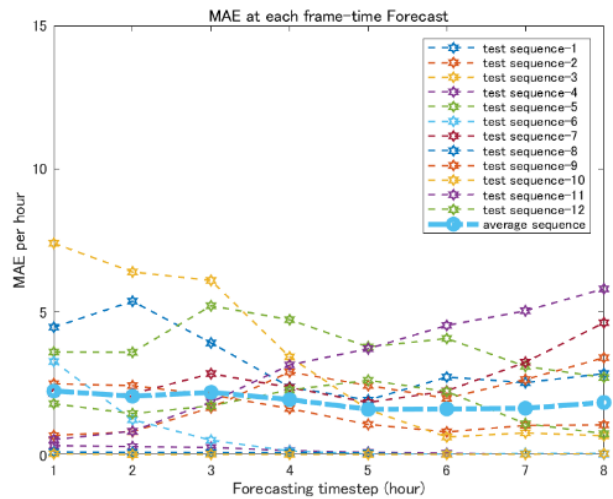

**Fig. 9.** Plot of MAE hourly accuracy of 1- and 2-time step forecasting tests based on 2×2frames 4 hours rain-code.

### 4.3 Rain-Code Based Forecasting Results and Accuracy Using 3×4 frames with Padding 1×4 mask

Figure. 10 shows five test sets of prediction results at each line that contains the "ground truth" of rain-code (the left-side of two images) and the 2 time step rain-code based "forecasting result" (the right-side of two images), where each image is a rain-code feature composed by 12 hours range of 3×4 frames. Though this initial frame size is rectangle, but we added the padding 1×4 mask at the bottom in order to keep the ratio of row-column to be the square shape.

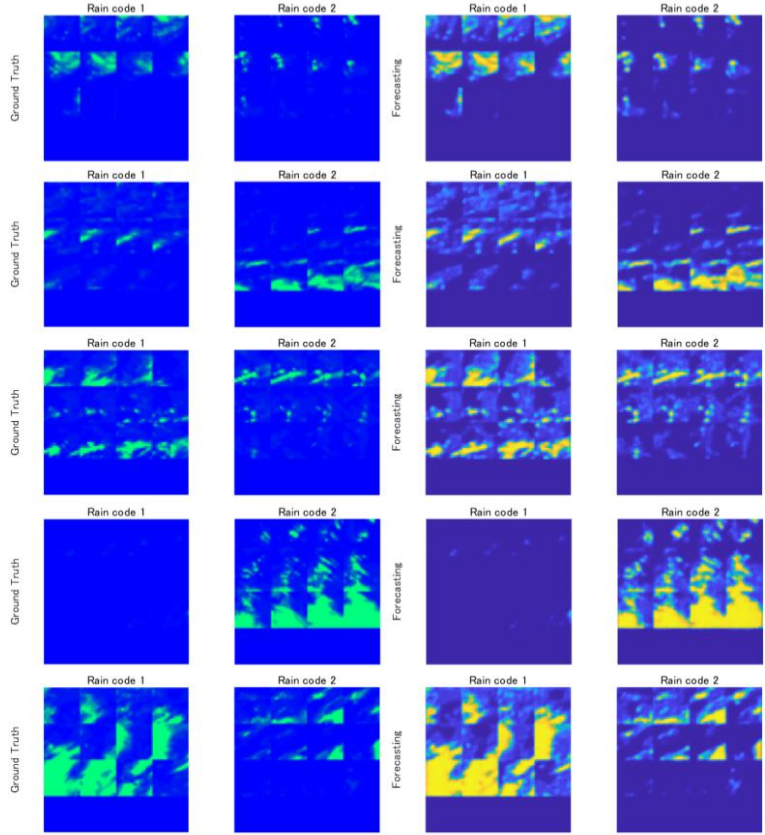

**Fig. 10.** Ground truth (left-side of two images) and 2 time step forecasting results (right-side of two images) using 3×4frames 12 hours rain-code with padding 1×4 mask at the bottom line.

Figure. 11 shows a Plot of root mean square error (RMSE) of 1- and 2-time step rain-code based forecasting test results. Here, the number of test rain-code with 3×4 frame is 12 cases. Each RMSE is computed into the range from 0 to 30, and the average RMSE among 12 cases is around 10 level. In addition, Figure. 12 shows a Plot of mean absolute error (MAE) of 1- and 2-time step rain-code based forecasting test results. Each MAE is computed into the range from 0 to 20, and the average MAE among 12 cases is around 5 level.

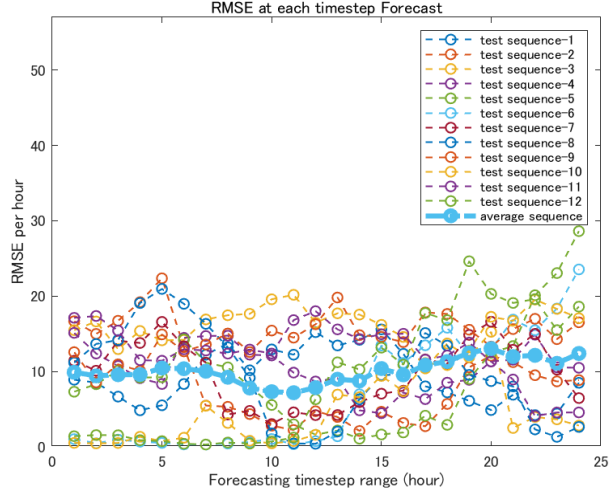

**Fig. 11.** Plot of RMSE hourly accuracy of 1- and 2-time step forecasting tests based on 3×4frames 12 hours rain-code with padding 1×4 mask.

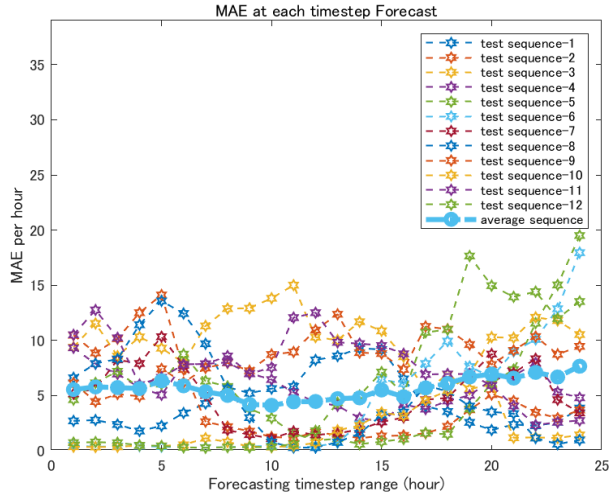

**Fig. 12.** Plot of MAE hourly accuracy of 1- and 2-time step forecasting tests based on 3×4frames 12 hours rain-code with padding 1×4 mask.

## 5    Concluding Remarks

### 5.1   Multi-Frame Based Code-to-code Spatiotemporal Forecasting

This paper proposed a rain-code based approach for spatiotemporal precipitation forecasting. We propose a novel rainy feature fusion that represents an unified multi-frame spatiotemporal relation for seemingly forecasting timestep reduction. We performed

14rain-code studies with various number of multi-frame using the algorithm of Code-to-code ConvLSTM for more than 6-hours forward precipitation forecasting. We applied our method to a dam region focusing on the Japanese rainy term May to October approximately 37 thousands hourly precipitation data since 2006 to 2019. We used the radar analysis hourly data for the central broader region with an area of $136 \times 148$ km$^2$. Although the forecasting results were slightly blurred and within a region narrower than the ground-truth region, the rainy locations of precipitation has been accurate. We presented several training results and prediction error indices with various number of multi-frame of the rain-code. Compared with four cases, the size of $2 \times 2$ frames rain-code outperform accuracy indices with both RMSE and MAE. The rain-code based approach allowed to keep the hourly average accuracy almost constant level at precipitation forecasting using our proposed Code-to-code ConvLSTM.

### 5.2 Future Extend Forecasting Range for Dam Inflow Prediction

In this study, the ConvLSTM algorithm was used for rain-code based spatiotemporal forecasting. Although convolutional LSTM has limitations with regard to the forecasting timestep, we can perform two-timestep forecasting using the 2-layer ConvLSTM network. We will tackle to predict the dam inflow toward high water under heavy rain. As the input for dam inflow model, we will utilise the 6-hours-forward precipitation for predicting the target dam inflow variable. This study focused on the rain-code of the multi-frame precipitation feature, but there are another important weather fusion, e.g., the sea surface temperature. Such weather movie data are usable for multi-frame-based spatiotemporal forecasting. Furthermore, with on-the-ground digital sensing, e.g., of the river water level at many locations, when we can set many sensors in multiple sub-region and temporally monitor condition, we believe that multi-frame based method allows the matrix of on-the-ground intensity codes to be used to extend the forecasting range by employing the code-to-code convolutional LSTM architecture.

**Acknowledgements.** We gratefully acknowledge the help provided by constructive comments of the anonymous referees. We thank Takuji Fukumoto and Shinichi Kuramoto (MathWorks Japan) for providing us with MATLAB resources.

## References

1. Ministry of Land Infrastructure, Transport and Tourism : Basic policy strengthening flood control function of existing dam, December 12 2019.
2. Sit M, Demiray B, Xiang Z et al : A Comprehensive Review of Deep Learning Applications in Hydrology and Water Resources, arXive:2007.12269, 2020.
3. Wu H, Yang Q, Liu J et al : A Spatiotemporal Deep Fusion Model for Merging Satellite and Gauge Precipitation in China, Journal of Hydrology, 584, pp124664, 2020.
4. Yan Q, Ji F, Miao K et al : Convolutional Residual Attention: A Deep Learning Approach for Precipitation Nowcasting, Advances in Meteorology, 2020.
5. Chen L, Cao Y, Ma L et al : A Deep Learning-Based Methodology for Precipitation Nowcasting With Radar, Earth and Space Science, 7(2), p.e2019EA000812, 2020.

6. Shi X., Yeung D.-Y. : Machine Learning for Spatiotemporal Sequence Forecasting: Survey, *arXiv:1808.06865v1*, 2018.
7. Shi X., Chen Z., Wang H., Yeung D.-Y., Wong W.-K., Woo W.-C. : Convolutional LSTM Network: A Machine Learning Approach for Precipitation Nowcasting, *arXiv:1506.04214v2*, 2015.
8. Shi X., Gao Z., Lausen L. et al. : Deep Learning for Precipitation Nowcasting: A Benchmark and A New Model, *NIPS*, 2017.
9. Wang Y., Long M., Wang J. et al : PredRNN: Recurrent Neural Networks for Predictive Learning using Spatiotemporal LSTMs, *NIPS*, 2017.
10. Wang Y., Gao Z. Long M. et al. : PredRNN++: Towards A Resolution of the Deep-in-time Dilemma in Spatiotemporal Predictive Learning, *ICML*, 2018.
11. Alleon A., Jauvion G., Quennenhen B et al. : PlumeNet: Large-Scale Air Quality Forecasting Using A Convolutional LSTM Network, *arXiv:2006.09204v1*, 2020.
12. Liu L., Zhang R., Peng J. et al. : Attentive Crowd Flow Machines, *arXiv:1809.00101v1*, 2018.
13. Wang J., Zhu W., Sun Y et al. : An Effective Dynamic Spatio-temporal Framework with Multi-Source Information for Traffic Prediction, *arXiv:2005.05128v1*, 2000.
14. Kim S, Hong S., Joh M et all : DeepRain: ConvLSTM Network for Precipitqation Prediction Using Multichannel Radar Data, $7^{th}$ *International Workshop on Climate Infomatics*, 2017.
15. Pascanu R., Mikolov T., Bengio Y. : On the Difficulty of Training Recurrent Neural Networks, *arXiv:1211.5063v2*, 2013.
16. Rumelhart D.E., Hinton G.E., Williams R.J. : Learning Representations by Backpropagating Errors, *Nature*, 323(6088), 533-536, 1986.
17. Graves A. : Generating Sequences with Recurrent Neural Networks, *arXiv:1308.0850v5*, 2014.
18. Gangopadhyay T., Tan S.-Y., Huang G. et al : Temporal Attention and Stacked LSTMs for Multivariate Time Series Prediction, NIPS, 2018.
19. Srivastava N., Mansimov E., Salakhutdinov R. : Unsupervised Learning of Video Representations using LSTMs, *arXiv:1502.04681v3*, 2016.
20. Japan Meteorological Agency : Radar/Rain gauge-Analysed Precipitation data, web accessed at 30 September 2020, https://www.jma.go.jp/en/kaikotan/.
21. Rajalingappaa S. : Deep Learning for Computer Vision – Expert Techniques to Train Advanced Neural Networks using TensorFlow and Keras, Packt Publishing, 2018.
Francois C. : Deep Learning with Python, Manning Publications, 2018.